\pdfoutput=1

\documentclass[11pt]{article}

\usepackage[preprint]{acl}

\usepackage{times}
\usepackage{latexsym}
\usepackage{amsmath}
\usepackage{multirow}
\usepackage{booktabs}
\usepackage{arydshln}
\usepackage{makecell}
\usepackage{tcolorbox}
\usepackage[capitalize,noabbrev]{cleveref}
\usepackage{placeins}
\usepackage{dblfloatfix}
\usepackage{verbatim}
\usepackage{enumitem}

\usepackage[T1]{fontenc}

\usepackage[utf8]{inputenc}

\usepackage{microtype}

\usepackage{inconsolata}

\usepackage{graphicx}

\title{Knowledge-Instruct: \\Effective Continual Pre-training from Limited Data using Instructions}

\author{
    \makebox[.35\linewidth]{Oded Ovadia
    \thanks{Equal contribution. Corresponding author: odedovadia@microsoft.com}} \hspace{0.5cm} 
    \makebox[.35\linewidth]{{Meni Brief}\footnotemark[1]} 
    \AND
    \makebox[.35\linewidth]{Rachel Lemberg}\hspace{0.5cm}
    \makebox[.35\linewidth]{Eitam Sheetrit} 
    \\\\
    \makebox[\linewidth]{Microsoft Industry AI} 
}

\begin{document}
\maketitle
\begin{abstract}
While Large Language Models (LLMs) acquire vast knowledge during pre-training, they often lack domain-specific, new, or niche information. Continual pre-training (CPT) attempts to address this gap but suffers from catastrophic forgetting and inefficiencies in low-data regimes. We introduce \textbf{Knowledge-Instruct}, a novel approach to efficiently inject knowledge from limited corpora through pure instruction-tuning. By generating information-dense synthetic instruction data, it effectively integrates new knowledge while preserving general reasoning and instruction-following abilities. Knowledge-Instruct demonstrates superior factual memorization, minimizes catastrophic forgetting, and remains scalable by leveraging synthetic data from relatively small language models. Additionally, it enhances contextual understanding, including complex multi-hop reasoning, facilitating integration with retrieval systems. We validate its effectiveness across diverse benchmarks, including \textit{Companies}, a new dataset that we release to measure knowledge injection capabilities.
\end{abstract}



\begin{figure*}
  \centering
  \includegraphics[width=0.6\textwidth]{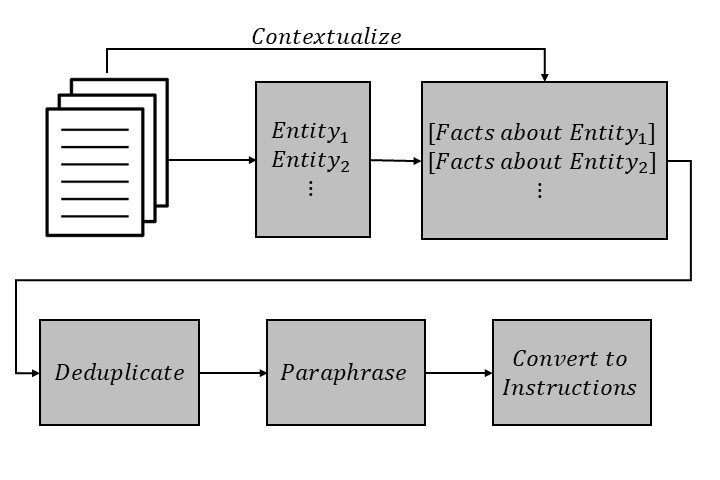}
  \caption{Visualization of the Knowledge-Instruct framework. A small text corpus is transformed into a set of information-dense instructions following the steps outlined in \cref{subsec:methodology}.}
\label{fig:knowledge_instruct}
\end{figure*}

\section{Introduction}\label{introduction}
Large language models (LLMs) encapsulate extensive knowledge within their pre-trained weights \citep{petroni2019language,cohen2023crawling,hu2023survey,chang2023survey}. This capacity is evidenced by the exceptional performance of modern LLMs in knowledge-intensive tasks \citep{achiam2023gpt,dubey2024llama,anthropic2024claude,team2023gemini,liu2024deepseek}. 

Such knowledge is predominantly acquired during the pre-training phase \citep{zhou2024lima}, which relies on extensive datasets containing trillions of tokens sourced from a wide array of domains, with a significant proportion derived from the web \citep{conneau2019unsupervised,gao2020pile,weber2024redpajama,touvron2023llama}. Many of these datasets, being representative of available textual content, include substantial amounts of duplicated material \citep{lee2021deduplicating,penedo2024fineweb}. Consequently, during the pre-training process, LLMs are trained on an enormous volume of factual information shaped by the underlying distribution of the training data.

The inherent repetition of knowledge within these datasets can significantly enhance LLM knowledge acquisition, as LLMs benefit from encountering different formulations of the same fact to thoroughly internalize and utilize it \citep{allen2023physics}. One example of this phenomenon is the so-called \textit{reversal curse} \citep{berglund2023reversal}, where LLMs trained on statements like "A is B" struggle to correctly learn the inverse relationship, "B is A." The abundance of related content within training datasets can help mitigate such issues, facilitating more robust knowledge acquisition.

One limitation of this heavy reliance on general text corpora is that many less common pieces of factual information are significantly underrepresented in many training datasets, making them less likely to be effectively learned by the model \citep{kandpal2023large}. Examples include domain-specific knowledge, geographically localized content, or other niche information. Moreover, certain data may be entirely absent from the training corpus, such as events that occurred after the model's knowledge cutoff or proprietary datasets unavailable during pre-training.

So, teaching LLMs niche knowledge is challenging due to the limited variations in fact representations within data-constrained corpora. This highlights a fundamental limitation in many practical scenarios. For instance, a manual, handbook, or textbook might comprehensively cover a topic using only $\approx 100$K tokens with minimal repetition. While this approach is highly effective for human learners, it poses significant challenges for LLMs, which rely on diverse and repeated formulations to effectively internalize information.

\textbf{Knowledge-Instruct:} In this work, we introduce \textit{Knowledge-Instruct}, a novel approach for teaching pre-trained LLMs new knowledge from small corpora using synthetic instruction data. This method enables efficient knowledge acquisition by directly addressing the challenges outlined earlier.

\noindent Knowledge-Instruct offers several key advantages over existing alternatives:

\begin{enumerate}[leftmargin=1pt]
    \item \textbf{Superior Factual Memorization:} Effectively learns factual information, outperforming other CPT methods.
    \item \textbf{Compatibility with Instruct Models:} Facilitates continual pre-training directly on instruction-tuned models, reducing reliance on unsupervised training that can disrupt chat templates and is often restricted for API-based models.
    \item \textbf{Minimizes Catastrophic Forgetting:} Exhibits minimal degradation in other model capabilities while integrating new knowledge.
    \item \textbf{Cost-Efficient:} Requires only a relatively small language model to generate synthetic training data.
    \item \textbf{Enhanced Context Understanding:} Strengthens the model’s ability to interpret and reason over retrieved context more effectively, making retrieval-augmented systems more accurate and reliable, even in challenging multi-hop scenarios.
\end{enumerate}

To validate these advantages, we empirically evaluate Knowledge-Instruct across diverse datasets and models. Our empirical results confirm its effectiveness in efficiently integrating new knowledge while preserving model stability and conversational fluency. Notably, our method achieves strong performance even with limited data, making it particularly suited for long-tail knowledge and domain-specific applications.

\section{Related Work}\label{related_work}

\paragraph{Continual Pre-training}
Continual pre-training adapts pre-trained language models to specialized domains through additional training on domain-specific corpora. Prior work has demonstrated success in broad domains like medicine \citep{chen2023meditron, wu2023pmcllamabuildingopensourcelanguage}, code \citep{rozière2024codellamaopenfoundation}, and mathematics \citep{shao2024deepseekmath, mitra2024orcamathunlockingpotentialslms} using massive datasets (10B–500B tokens). However, these methods fail to scale to small corpora (\textasciitilde1M tokens), where limited diversity and repetition hinder effective knowledge acquisition \citep{kandpal2023large}. This challenge stems from the inherent data inefficiency of language models, which require multiple contextual exposures to internalize facts \citep{allen2023physics}. Recent approaches address this through synthetic data augmentation, generating paraphrased texts \citep{maini2024rephrasing,ovadia2023fine} or structured entity-based expansions \citep{yang2024synthetic}. While \citet{yang2024synthetic}'s method is highly effective for adding new knowledge, it demands a very large amount of tokens, and also requires an additional instruction-tuning phase to be useful in practice.

\paragraph{Knowledge Injection} 
Alternative methods for knowledge integration include \textit{knowledge editing}, which updates specific model parameters to insert facts while preserving existing capabilities \citep{mitchell2021fast,meng2022mass}. While effective for atomic edits, these methods struggle to scale for broader knowledge. Retrieval-augmented generation (RAG) \citep{lewis2020retrieval} bypasses parametric learning by dynamically accessing external documents, but depends on retrieval quality, and potentially demands many tokens per query. Synthetic data generation allows for creating diverse synthetic pre-training corpora via complex external prompting ~\citep{li2023textbooks, mukherjee2023orcaprogressivelearningcomplex}, combined with further training. Supervised fine-tuning (SFT), particularly instruction tuning \citep{wang2022supernaturalinstructionsgeneralizationdeclarativeinstructions}, is highly effective at improving zero-shot and reasoning capabilities. However, SFT is primarily used to enhance task generalization rather than introducing new factual knowledge \citep{mitra2023orca,chia2023instructeval,zhou2024lima}.


\section{Knowledge-Instruct}\label{sec:methodology} 
The core idea behind this methodology is to transform a corpus of raw textual data into an information-dense set of factual instructions, thereby facilitating a knowledge-focused SFT phase.

\subsection{Rationale}\label{subsec:rationale} Our approach is guided by key observations and hypotheses that we see as fundamental to building knowledge injection framework for LLMs:


\begin{itemize}[leftmargin=1pt,itemsep=2pt]
    \item \textbf{Entities:} Knowledge is intrinsically linked to entities, serving as primary anchors for factual information. 
    \item \textbf{Coverage:} A training dataset must thoroughly capture all pertinent factual details about the targeted entities in order to successfully integrate new knowledge.
    \item \textbf{Context:} The meaning and representation of knowledge often depend on surrounding text, making clear and complete semantic context essential. 
    \item \textbf{Repetition:} Each piece of factual information should be presented multiple times (e.g., through paraphrasing) for more robust learning. 
    \item \textbf{Knowledge Distribution:} The dataset’s distribution of factual information should closely reflect the patterns found in the original corpus. 
\end{itemize}


\noindent We accomplish these objectives through a six-step process consisting of entity extraction, factual extraction, contextualization, deduplication, paraphrasing, and instruction conversion. A visual overview of these steps is depicted in ~\cref{fig:knowledge_instruct}.

\subsection{Methodology}\label{subsec:methodology}

Given a corpus $\mathcal{D} = \{d_1, d_2, \dots, d_N\}$ of $N$ documents, a pre-trained language model $\mathcal{M}_{ext}$ for extraction, we construct a training dataset $\mathcal{D}_{\text{train}}$ in six stages:

\paragraph{(1) Entity Extraction.}
For each document $d \in \mathcal{D}$, extract the set of entities $\mathcal{E}_{d,\mathcal{M}_{ext}}$ that appear in $d$, as identified by $\mathcal{M}_{ext}$:
\begin{equation*}
    \mathcal{E}_{d,\mathcal{M}_{ext}} \;=\; \{\, e \mid e \; \text{identified by } \mathcal{M}_{ext} \text{ in } d \}.
\end{equation*}

\paragraph{(2) Factual Extraction.}
Using the entities $\mathcal{E}_{d,\mathcal{M}_{ext}}$, extract all facts $f = (e, \text{detail about} \; e)$, where $e\in 
\mathcal{E}_{d,\mathcal{M}_{ext}}$. Concretely,
\begin{equation*}
    \mathcal{F}_{d,\mathcal{M}_{ext}} = \{ f \mid \mathcal{M}_{ext} \text{ extracted } f \text{ from } d\text{ about }e\},
\end{equation*}
where $\mathcal{F}_{d,\mathcal{M}_{ext}}$ denotes the set of facts found in the document $d$. This process is repeated several times, each time prompting $\mathcal{M}_{ext}$ to search for missing facts and maximize coverage.

\paragraph{(3) Contextualization.}
For each fact $f \in F_{d,\mathcal{M}_{ext}}$, we use $\mathcal{M}_{ext}$ to ensure that $e$ appears explicitly within $f$, thus clarifying the context of $f$. The context-augmented fact is thus $\bigl(f, C_{f}\bigr) $.

\paragraph{(4) Deduplication.}
Within each set $F_{d,\mathcal{M}_{ext}}$, identify and remove equivalent facts using $\mathcal{M}_{ext}$. We denote the deduplicated set as $F_{d,\mathcal{M}_{ext}}^*$.

\begin{table*}[!t]
    \centering
    \footnotesize 
    \renewcommand{\arraystretch}{1.3} 
    \setlength{\tabcolsep}{6pt} 
    \begin{tabular}{l | c c c c | c}
        \toprule
        \textbf{Model} & \textbf{Companies} & \textbf{PopQA} & \textbf{MultiHopRAG} & \textbf{MultiHopRAG + Oracle} & \textbf{Average} \\
        \midrule 
        \multicolumn{6}{c}{\textbf{Llama-3.1-8B-Instruct}} \\
        \midrule
        Base                 & 0.0  & 13.2  & 22.4  & 70.7  & 26.6 \\
        CPT                  & 4.2  & 40.5  & 45.0  & 66.1  & 38.9 \\
        Rephrase CPT         & 17.7 & 49.2  & 41.2  & 66.7  & 43.7 \\
        Synthetic CPT        & 53.5 & 73.4  & 50.6  & 69.0  & 61.6 \\
        \textbf{Knowledge-Instruct}   & \textbf{81.8} & \textbf{76.8}  & \textbf{56.5}  & \textbf{80.0}  & \textbf{73.6} \\
        \midrule
        \addlinespace[5pt]
        \multicolumn{6}{c}{\textbf{Phi-4-14B}} \\
        \midrule
        Base                 & 1.2  & 5.0   & 26.5  & 74.0  & 26.7 \\
        CPT                  & 9.4  & 31.8  & 33.4  & 75.3  & 37.5 \\
        Rephrase CPT         & 13.3 & 32.7  & 36.7  & 75.6  & 39.6 \\
        \textbf{Knowledge-Instruct}   & \textbf{86.2} & \textbf{80.1}  & \textbf{60.9}  & \textbf{79.6}  & \textbf{76.7} \\
        \midrule
        \addlinespace[5pt]
        \multicolumn{6}{c}{\textbf{GPT-4o}} \\
        \midrule
        Base               & 4.8  & 62.4  & 45.6  & 82.6  & 48.9 \\
        \bottomrule
    \end{tabular}
    \caption{Main results comparing Knowledge-Instruct with other methods. "Base" refers to the original model, without any training, and "MultiHopRAG + Oracle" refers to direct access to the relevant ground-truth documents.}
    \label{tab:main_results}
\end{table*}

\paragraph{(5) Paraphrasing.}
For each fact $f \in F_{d,\mathcal{M}_{ext}}^*$, generate a set of $k$ paraphrased variations $\mathcal{P}_f$ using $\mathcal{M}_{ext}$:
\begin{equation*}
    \mathcal{P}_f \;=\; \{\, p_1, \dots, p_k \mid p_i \equiv f \}.
\end{equation*}

\paragraph{(6) Instruction Conversion.}\label{par:instruction_conversion}
Finally, convert each paraphrased fact $p \in \mathcal{P}_f$ into an instruction-response pair $I_p = \bigl(\text{instruction}, p\bigr)$, 
where \textit{instruction} is a prompt that $p$ is a valid response for. This step is necessary to ensure suitability for supervised fine-tuning (SFT).

This conversion is performed using a simple rule-based approach, where facts are mapped to instruction-response pairs through pre-defined templates. These templates include variations such as:
\begin{quote}
     \textit{"Tell me a fact about \{entity\}."}\\
    \textit{"What can you tell me about \{entity\}?"}\\
\textit{    "Please provide a fact about \{entity\}."}   
\end{quote}


\noindent This ensures diversity in instruction phrasing while maintaining consistency in knowledge representation. The full list of templates is shown in \cref{app:sft_conversion}.

\paragraph{(7) Putting it all together.}
The final dataset $\mathcal{D}_{\text{train}}$ is the aggregation of all instruction-response pairs generated:

\begin{equation*}
    \mathcal{D}_{\text{train}} \;=\; \bigcup_{d \in \mathcal{D}}\, \bigcup_{f \in F_{d,\mathcal{M}_{ext}}} \, \bigcup_{p \in \mathcal{P}_f} I_p.
\end{equation*}

It is worth noting that entity and factual extraction are similar to constructing a knowledge graph. A knowledge graph is a directed graph $G = (V, E)$, where  $V = \{v_1, \dots, v_{|V|}\}$ represents the set of all possible entities, 
and $E \subseteq V \times R \times V$ represents the set of all facts or factual relations involving these entities. While we assume that $\mathcal{E}_{d,\mathcal{M}_{ext}} \approx V$, it is likely that $\mathcal{F}_{d,\mathcal{M}_{ext}}^* \subset E$ as we do not explicitly attempt to construct or traverse a graph.

\begin{table*}[h]
    \centering
    \small
    \renewcommand{\arraystretch}{1.2} 
    \setlength{\tabcolsep}{5pt} 
    \begin{tabular}{l c c c c c c}
        \toprule
        & \textbf{Knowledge-Instruct} & \textbf{Synthetic CPT} & \textbf{CPT} & \textbf{Rephrase CPT} & \textbf{Base} & \textbf{Base + Orca SFT} \\
        \midrule
        MMLU          & 66.7 & 67.0 & 66.9 & 66.6 & \textbf{68.2} & 67.3 \\
        TriviaQA      & 67.0 & 67.8 & 68.0 & 67.1 & \textbf{70.1} & 69.0 \\
        ARC-Challenge & \textbf{58.2} & 53.4 & 56.3 & 56.3 & 50.3 & 56.9 \\
        ARC-Easy      & \textbf{83.8} & 80.6 & 82.2 & 82.7 & 75.8 & 83.3 \\
        GSM8K         & 75.7 & 67.1 & 65.5 & 69.6 & \textbf{77.6} & 75.6 \\
        Hellaswag     & 76.3 & \textbf{77.1} & 75.7 & 73.2 & 70.0 & 72.6 \\
        Winogrande    & \textbf{68.0} & 65.9 & 65.1 & 66.2 & 65.4 & 67.2 \\
        OpenBookQA    & 44.4 & \textbf{44.8} & 41.4 & 41.4 & 38.4 & 42.6 \\
        MMLU-Pro      & 35.3 & 29.8 & 22.6 & 22.1 & 33.6 & \textbf{38.2} \\
        \midrule
        Average       & \textbf{63.9} & 61.5 & 60.4 & 60.6 & 61.0 & 63.6 \\
        \bottomrule
    \end{tabular}
    \caption{Performance comparison of different methods on various general LLM benchmarks. All results are reported for Llama trained on the MultiHop-RAG data. "Base" refers to the original model without any additional training, while "Base + Orca SFT" denotes the model fine-tuned exclusively on OpenOrca data, serving as baselines for comparison. Evaluation was conducted using the LM-Evaluation-Harness framework \cite{eval-harness}.}
    \label{tab:catastrophic_benchmark_results}
\end{table*}


\section{Experiment Setup}
\subsection{Datasets}\label{datasets}
To properly evaluate the proposed method, we focus on datasets where the model's prior knowledge is either limited or nonexistent. We select three diverse datasets: Companies, PopQA, and MultiHop-RAG.

Each dataset represents a distinct type of knowledge-intensive challenge: Companies contains entirely unseen knowledge, making it a true test of the model's ability to learn new facts. PopQA, on the other hand, includes knowledge that the model has likely encountered during pre-training but in a long-tail distribution, assessing its ability to retrieve and generalize from previously seen but less common data. Finally, MultiHop-RAG requires reasoning over a complex knowledge graph, testing the model’s ability to synthesize multiple pieces of information and derive meaningful insights from them. For all datasets we use open-ended questions only, to avoid any guessing that multiple choice questions may enable. The code for creating Companies, as well as the full dataset and PopQA subset are made publicly available\footnote{\url{https://github.com/meniData1/knowledge-instruct}}.

\textbf{Companies:} We created a synthetic dataset composed of 23 entirely fictional companies. To do so, we first created a set of unrealistic company names alongside a very short description of the company. We then used GPT-4o~\citep{OpenAI2024} to generate an overview describing the company, a full catalog of the products/services they offer, and their financial report for a recent quarter. The prompts for this can be found in ~\cref{app:prompts}. A full example from the dataset can be found in~\cref{app:data_samples}. For each section, for each company, we generated 10 questions, resulting in a total of 690 questions.

\textbf{PopQA:} PopQA~\citep{mallen2022not} is a dataset containing long-tail knowledge gathered from Wikipedia. Long-tail knowledge in this context refers to the popularity of a page, measured as monthly user visits. PopQA questions have data about the popularity of both the 'Subject' of the question and the 'Object', or answer, to the question. We selected a subset of PopQA such that:
$\text{Popularity}_{\text{Subject}} + \text{Popularity}_{\text{Object}} < 2500$,  thus ensuring the long-tail property. Next, we removed all ambiguous and problematic questions, by keeping only questions that GPT-4o was able to answer correctly when provided with the relevant Wikipedia article. This resulted in a total of 1,935 question and answer pairs, alongside their respective Wikipedia articles. 

\textbf{MultiHop-RAG}: MultiHop-RAG \citep{tang2024multihop} is a dataset designed to evaluate RAG systems on complex multi-hop queries, where answering a question requires reasoning across multiple documents to connect disparate pieces of information. It comprises a knowledge base of English news articles, a diverse set of multi-hop questions, their corresponding ground-truth answers, and the necessary supporting evidence. The dataset was constructed by extracting factual sentences from news articles, rephrasing them into claims, and generating multi-hop queries that require reasoning across multiple documents.

\subsection{Evaluation}
To simulate real-life applications, all questions are in conversational open-ended format. So, standard accuracy metrics were not applicable. Instead, we employed an LLM-as-a-Judge~\citep{Zheng2023} for evaluation. Specifically, we used the judge suggested by~\citet{brief2024mixing}, with GPT-4o as the LLM and their suggested prompt. A response is considered correct only if it strictly matches (i.e., receiving a score of 2 out of \{0,1,2\}), with normalization performed by dividing the score by two.

\section{Experiments}\label{results}

We conduct a thorough exploration of the capabilities of the proposed method on the datasets described in \cref{datasets}. We apply Knowledge-Instruct on two different language models: Llama-3.1-8B-Instruct (Llama) \cite{dubey2024llama} and Phi-4-14B ~\citep{abdin2024phi}. 

\subsection{Baselines}
We compare Knowledge-Instruct to several methods, described below. Unless otherwise stated, all data was generated using GPT-4o-mini~\citep{openai_gpt4o_mini}.

\noindent\textbf{Continual Pre-training (CPT):} We perform the standard CPT approach, in which we directly train the LLM on the raw text corpora in an unsupervised manner. However, this process breaks the chat template of the model, and results in a model that has very poor conversational and instruction-following abilities. To mitigate that, we perform another step of SFT on $10,000$ high-quality instructions taken from OpenOrca ~\citep{OpenOrca}.

\noindent\textbf{Rephrase CPT:} Following ~\citep{allen2023physics,maini2024rephrasing,ovadia2023fine}, we also test a different CPT approach, where each document is rephrased in several different styles. Specifically, we rephrase each document $100$ times, using $10$ different prompts to increase diversity. The full list of prompts is detailed in~\cref{app:prompts}. Then, we perform the exact same CPT process described before on the paraphrased corpora. 

\noindent\textbf{Synthetic Continued Pre-training}: We used the methodology proposed by~\citet{yang2024synthetic} to create the Synthetic CPT models. Due to its relatively small size we created entity pairs and triplets for the companies dataset, while for the larger PopQA and MultiHopRAG datasets we used only entity pairs. Due to similar computational constraints, we skipped Phi-4. Using their training method exactly, we first created a CPT-only model (with a mixture of general data), followed by instruction tuning.  

\subsection{Main Results}\label{subsec:main_results}

\begin{figure}[t]
  \includegraphics[width=\columnwidth]{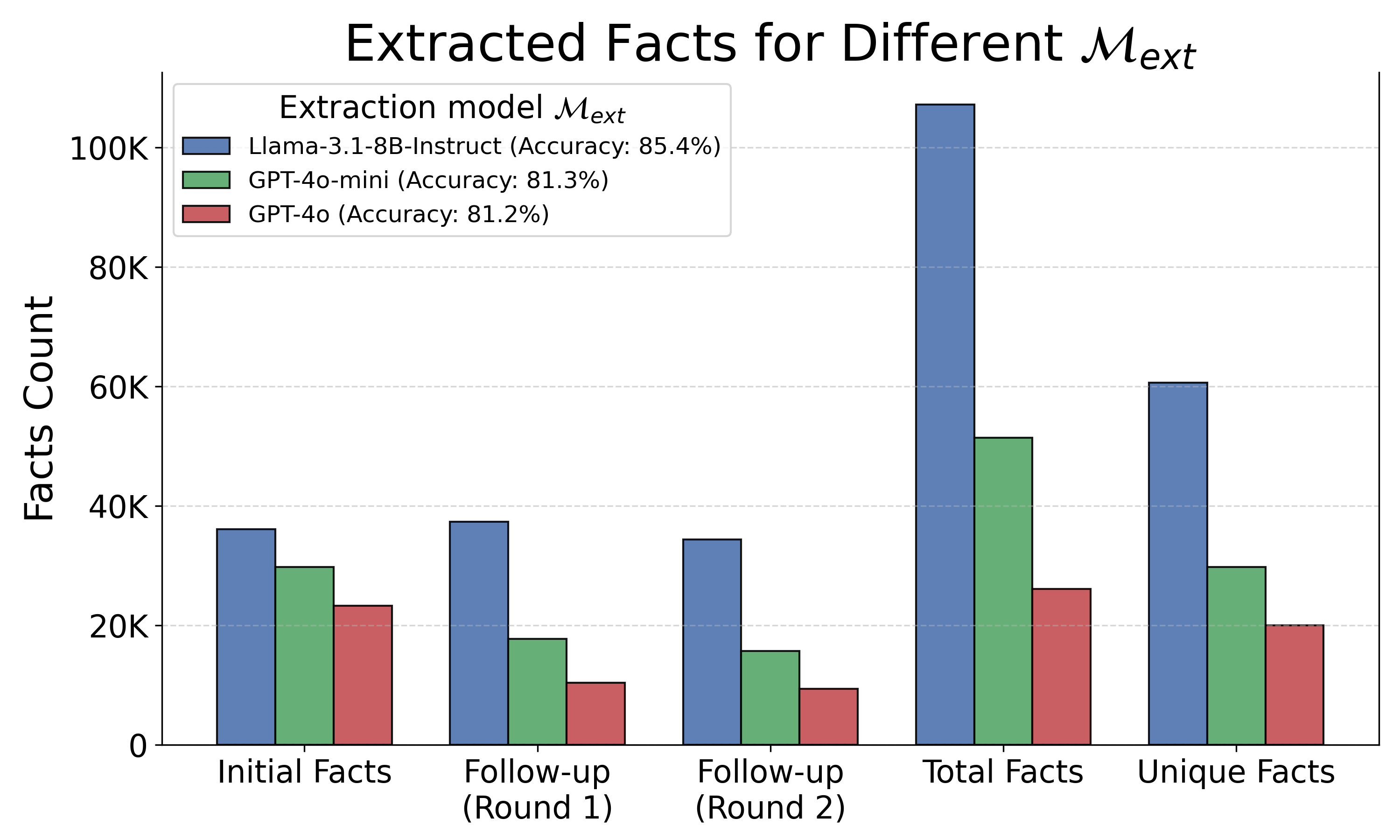}
  \caption{Number of facts extracted by different LLMs at various stages of the Knowledge-Instruct process. The extraction is performed in three rounds: an initial extraction followed by two iterative verification passes, where the model identifies any missed facts. The results are reported for the Companies dataset, with 'Total Facts' representing the cumulative count across all rounds and 'Unique Facts' indicating distinct, non-redundant extractions. The final model accuracy on the benchmark is provided in the legend.}
  \label{fig:facts_extraction}
\end{figure}


The complete evaluation results for all datasets and methods are presented in \cref{tab:main_results}. Across all cases, Knowledge-Instruct consistently achieves the best results, outperforming the other methods and significantly enhancing the models' knowledge. To better understand these outcomes, we analyze each dataset separately, highlighting key trends and differences in model performance.

\textbf{Companies:} This dataset introduces entirely new knowledge, as seen in the near-zero "Base" scores, confirming that the models have no prior exposure apart from random guesses. A clear trend emerges: CPT and Rephrase CPT \textit{fail completely} to acquire meaningful information, performing poorly on both Llama and Phi-4. While rephrasing provides a slight boost, scores remain low. Synthetic CPT improves over the other baselines but still somewhat struggles. 

\noindent In contrast, Knowledge-Instruct significantly outperforms all methods, exceeding 80\% accuracy for both models, demonstrating its ability to adapt to unseen data. It is slightly more effective for Phi-4, possibly due to its larger parameter count.

\textbf{PopQA:} Unlike Companies, PopQA contains knowledge that the models have likely encountered during pre-training, though infrequently. This is reflected in the higher "Base" scores and the better performance of CPT and Rephrase CPT, which - despite failing on Companies - show a notable improvement here. However, they still lag behind Synthetic CPT and Knowledge-Instruct, the latter once again achieving the best results.

\noindent Interestingly, Llama outperforms Phi-4 with CPT and Rephrase CPT. We hypothesize this may be due to Phi's stronger reliance on highly curated synthetic training~\citep{abdin2024phi}, leading to less exposure to raw Wikipedia content. Nonetheless, Knowledge-Instruct shows consistent gains across both models, following the same trend observed in Companies. Furthermore, both Knowledge-Instruct and Synthetic CPT outperform the much larger GPT-4o, which was likely exposed to this data before, suggesting that even large models struggle with long-tail knowledge~\citep{kandpal2023large}. 

\noindent \textbf{MultiHop-RAG}: This dataset presents a different challenge, requiring reasoning over retrieved knowledge rather than just factual recall. Even when the providing the model with the full correct context (we refer to this as the "Oracle" setting), GPT-4o reaches a ceiling of 82.6\%, indicating the difficulty of the task.

\noindent Similar to PopQA, prior exposure to similar training distribution appears to play a role in the standard case (without "Oracle"), with Llama achieving higher scores than Phi-4 with CPT and Rephrase CPT, likely due to the same factors discussed earlier. However, Rephrase CPT does not show a consistent improvement over CPT here.

\noindent Once again, Knowledge-Instruct achieves the best results, though overall scores remain lower than in the other two datasets, emphasizing the greater complexity of this benchmark. However, when adding the Oracle case, there is a significant improvement. A notable trend emerges: all CPT methods fail to significantly improve Oracle results, and in some cases, even degrade performance. Knowledge-Instruct, on the other hand, significantly improves results, particularly for Llama, which sees an almost 10\% gain. This indicates that Knowledge-Instruct not only preserves reasoning capabilities over retrieved data but actively enhances them, highlighting its strong potential for (RAG) applications.

\begin{figure}[t]
  \includegraphics[width=\columnwidth]{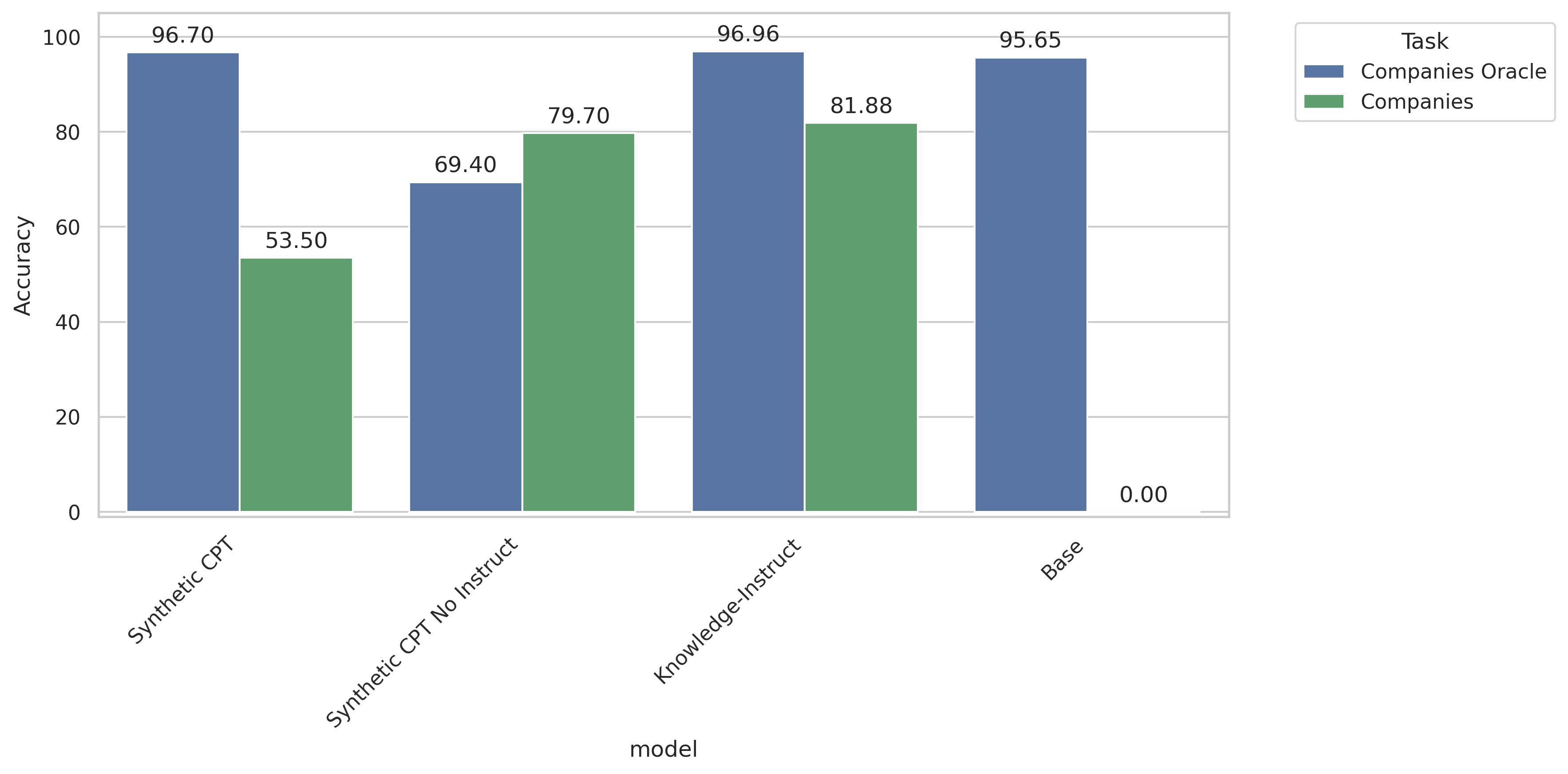}
  \caption{A comparison of Knowledge-Instruct with Synthetic CPT. Synthetic CPT without further SFT is better at the new domain, at the expense of instruction following, and vice versa. The base model, Llama, is shown for reference.}
  \label{fig:model_comparison}
\end{figure}

\subsection{Catastrophic Forgetting Analysis}\label{subsec:forgetting}
One major fine-tuning risk is the phenomenon of \textit{catastrophic forgetting}
\cite{kirkpatrick2017overcoming,goodfellow2013empirical,chen2020recall,luo2023empirical}, where models lose some of the capabilities they had prior to the fine-tuning process. 

To address this concern, we perform a case-study on Llama in the MultiHop-RAG scenario, and evaluate all of the fine-tuned models on a range of common general LLM benchmarks. This includes knowledge-intensive tasks (MMLU, TriviaQA~\citep{hendrycks2020measuring,joshi2017triviaqa}), reasoning tasks (ARC, GSM8K, Winogrande~\citep{Clark2018ThinkYH,cobbe2021training,sakaguchi2021winogrande}), reading comprehension (OpenBookQA~\citep{OpenBookQA2018}), and a combination of knowledge with reasoning (MMLU-Pro~\citep{wang2024mmlu}).

\textbf{General Results:} The full results are presented in \cref{tab:catastrophic_benchmark_results}. Overall, Knowledge-Instruct preserves general capabilities and mitigates catastrophic forgetting more effectively than other methods. In fact, we observe an improvement in reasoning tasks across all methods, likely due to the inclusion of high-quality SFT data. This aligns with the "Base + Orca SFT" column, where OpenOrca alone improves Llama’s reasoning ability. While all methods show minor deterioration in general knowledge tasks (MMLU and TriviaQA), they largely maintain or even slightly improve performance on the other benchmarks, likely due to the added supervised fine-tuning (SFT) data.

However, there are two notable exceptions. In GSM8K, Knowledge-Instruct shows only minimal deterioration, whereas other methods exhibit significant decline. Similarly, in MMLU-Pro Knowledge-Instruct retains capabilities, even slightly surpassing the base model, while the CPT methods show worse results.

\textbf{CPT vs. SFT:}
~\cref{fig:model_comparison} show the performance of Knowledge-Instruct compared to Synthetic CPT, before and after adding the SFT stage. All SFT models, including the base model that has no prior knowledge at all, achieve over 95\% accuracy in the Companies Oracle case. While the performance of Synthetic CPT before instruction tuning is significantly better on Companies directly, its under-performs on the much easier Oracle version. In contrast, Knowledge-Instruct performs well on both versions, showing full retention of previous abilities, with no knowledge learning trade-offs.

\textbf{Regularization Hypothesis:} We hypothesize that this is due to Knowledge-Instruct’s single-step approach, where general SFT data and new knowledge data are integrated simultaneously, likely serving as a form of regularization~\citep{brief2024mixing}. In contrast, CPT-based methods follow a two-step approach - an initial unsupervised pre-training phase, followed by SFT. Our findings, both in the Oracle cases, and for general benchmarks like GSMK8K or MMLU-Pro, suggest that without proper regularization, CPT-based methods impair the model’s ability to reason over existing knowledge.


\subsection{Effect of Paraphrasing}
We conduct an ablation study to assess the impact of the paraphrasing step in Knowledge-Instruct. Specifically, we generate facts as described in \cref{sec:methodology}, creating five paraphrases for each fact. We then fine-tune the LLM with progressively more paraphrases, starting from only the original fact and increasing up to the full set of five.

\begin{figure}[t]
  \includegraphics[width=\columnwidth]{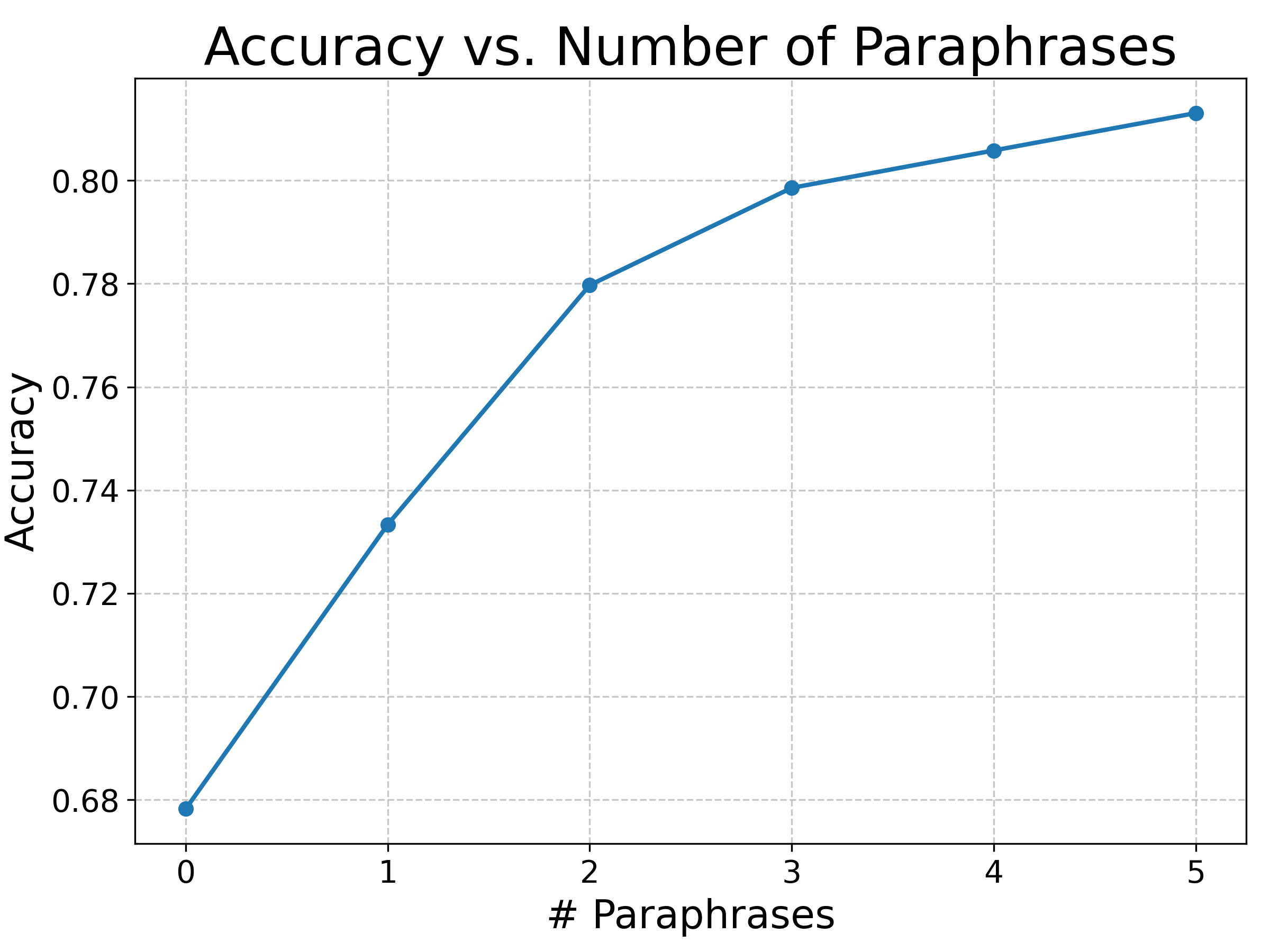}
  \caption{Effect of the paraphrasing step in Knowledge-Instruct on accuracy using the Companies dataset with Llama.}
  \label{fig:paraphrase_ablation}
\end{figure}

The results of this study on the Companies dataset using Llama are shown in \cref{fig:paraphrase_ablation}. We observe a clear trend: performance improves as the number of paraphrases increases, and begins to plateau around 3 paraphrases. This aligns with previous findings \citep{ovadia2023fine,allen2023physics}, which report similar benefits from diverse rephrasings. 

While a high number of paraphrases is generally beneficial, practical computational considerations must be taken into account. Notably, our results show that just five paraphrases are sufficient to yield a significant performance boost. This suggests that Knowledge-Instruct is already traversing the latent knowledge graph of the corpus effectively, requiring only a modest amount of paraphrasing to enhance learning.

\subsection{Extraction Model Ablation}

We perform another ablation study on the impact of the LLM used to extract the facts, $\mathcal{M}_{ext}$ as described in~\cref{subsec:methodology}. Throughout all experiments we used GPT-4o-mini, which showed excellent results while being very fast and cheap. To measure its performance, we repeated the entire Knowledge-Instruct methodology as described in~\cref{sec:methodology} for the Companies dataset, using GPT-4o, GPT-4o-mini, and Llama. 

The full results are presented in~\cref{fig:facts_extraction}. We observe that Llama generates the highest number of facts, followed by GPT-4o-mini, while GPT-4o produces the fewest. However, many of the facts from Llama are repetitive, as it struggles to extract distinct follow-up facts and often repeats previously generated ones. GPT-4o-mini exhibits the same behavior to a lesser extent, while GPT-4o is the most efficient in generating unique facts. This repetition introduces a natural paraphrasing effect, which ultimately benefits the fine-tuned model trained on the Llama generated dataset, as reflected in~\cref{fig:paraphrase_ablation}, while increasing the required training tokens. 

Our results demonstrate that massive LLMs are not necessary for Knowledge-Instruct, as even smaller models can improve cost-effectiveness without significant trade-offs.

\begin{table}[!t]
    \centering
    \footnotesize
    \renewcommand{\arraystretch}{1.3}
    \setlength{\tabcolsep}{6pt}
    \begin{tabular}{l | c c | c}
        \toprule
        \textbf{Model} & \textbf{Oracle} & \textbf{Reconstruct} & \textbf{Gap} \\
        \midrule 
        PopQA Oracle & 99.0 & 97.1 & 1.9 \\
        Companies Oracle & 98.4 & 95.5 & 2.9 \\
        \bottomrule
    \end{tabular}
    \caption{Comparison of oracle and reconstructed scores for GPT-4o on the PopQA and Companies datasets.}
    \label{tab:oracle_comparison}
\end{table}

\subsection{Dataset Coverage}
~\cref{tab:oracle_comparison} shows our evaluation of the \textbf{coverage} of the data generated using Knowledge-Instruct, i.e., $\mathcal{D}_{train}$\footnote{Coverage is trivial for a full corpus, since all information is contained within. The further removed a training dataset is from the underlying corpus, the less trivial this becomes.}. To do so, we use Oracle versions of PopQA and Companies, with the full document provided as context. We set the baseline performance for GPT-4o in this setup at 99\% and 98.4\%  accuracy respectively. Next, we concatenate all the facts generated from the document and use this concatenation as the context. The accuracy for this version is 97.1\% and 95.5\% respectively, meaning the average Accuracy Degradation is under 2.5\%, suggesting that the generated facts provide strong coverage of the dataset. A more formal definition of this metric can be found in~\cref{app:acc_degradation}.

\section{Conclusions}
In this work, we introduced Knowledge-Instruct, a novel instruction-based fine-tuning approach for efficient knowledge injection in LLMs. Our experiments highlight its advantages over traditional CPT, including superior factual memorization, reduced catastrophic forgetting, and improved contextual understanding, all while remaining cost-effective by leveraging smaller language models for synthetic data generation.

\section{Limitations}
Knowledge-Instruct directly addresses an important limitation in LLMs by successfully adapting to niche knowledge. Despite its advantages, Knowledge-Instruct has certain limitations. First, while synthetic instruction generation facilitates knowledge injection from limited data, its quality and coverage depend on the prompting strategy, requiring adjustments for different datasets. Second, although our method reduces catastrophic forgetting, it does not fully eliminate it, necessitating further research into long-term knowledge retention. Lastly, as with all machine learning approaches, hyperparameter selection influences performance, making careful optimization essential for specific use cases.

\bibliography{ref}

\newpage
\onecolumn

\appendix
\section{Training Details}
\label{app:hyperparameters}
For the CPT, Rephrase CPT, and Knowledge-Instruct experiments, we used a learning rate of of $1e-05$, a batch size of 4, a context length of $2,048$. For the SFT phase of the CPT and Rephrase CPT training runs we used a lower learning rate of $1e-06$ to attempt to reduce catastrophic forgetting. For the Synthetic CPT experiments, we used the exact same code and configurations as detailed in the official repository \footnote{\url{https://github.com/zitongyang/synthetic_continued_pretraining}}. Training runs were conducted on pairs of 2$\times$H100 or 4$\times$A100 NVIDIA GPU nodes, subject to availability. 

\section{Accuracy Degradation}
\label{app:acc_degradation}

\paragraph{Aggregated-Fact Equivalence.}
Let $d$ be an arbitrary document in the corpus $\mathcal{D}$, and let 
\begin{equation}  
\mathcal{F}(d) = \{\,f_1, f_2, \dots, f_m\,\}
\end{equation}

be the set of all deduplicated and contextualized facts extracted from $d$. Define the aggregated-fact representation $\mathcal{A}(d)$ as the concatenation of all these facts:
\begin{equation}
\mathcal{A}(d) = \sum\limits_{f \in \mathcal{F}(d)} f.
\end{equation}

We aim that for any question $q$ pertaining to the content of document $d$, the performance of the language model $\mathcal{M}$ when answering based on the aggregated-fact representation $\mathcal{A}(d)$ be identical to its performance when provided with the full document $d$. Formally, letting $P(q \mid x)$ denote the probability that $\mathcal{M}$ answers question $q$ correctly when conditioned on input $x$, our criterion is
\begin{equation}
    P\bigl(q \mid \mathcal{A}(d)\bigr) \;=\; P\bigl(q \mid d\bigr), \quad \forall \, q \in Q(d),
\end{equation}
where $Q(d)$ is the set of all questions about document $d$. In other words, when the aggregated facts are available, the model's performance on questions about the document should be equivalent to having access to the complete document.

\paragraph{Accuracy Degradation.}
Let $\mathcal{D}_{\text{target}}$ denote the set of all target documents. For each document $d \in \mathcal{D}_{\text{target}}$, define the degradation in accuracy $\Delta(d)$ as the summed difference in accuracy between answering questions based on the full document and answering based on the aggregated fact representation $\mathcal{A}(d)$:
\begin{equation}
    \Delta(d) \;=\; \sum_{q \in Q(d)} \Bigl( \; P\bigl(q \mid d\bigr) - P\bigl(q \mid \mathcal{A}(d)\bigr) \Bigr),
\end{equation}
where $Q(d)$ is the set of all questions that can be answered with full access to document $d$, and $P(q \mid x)$ denotes the accuracy (i.e., the probability of answering question $q$ correctly) when conditioned on input $x$.

The overall accuracy degradation metric for the dataset is then defined as the average of these degradations across all target documents:
\begin{equation}
    \Delta\bigl(\mathcal{D}_{\text{target}}\bigr) \;=\; \frac{1}{|\mathcal{D}_{target}|}\sum_{d \in \mathcal{D}_{\text{target}}} \Delta(d).
\end{equation}
A lower value of $\Delta\bigl(\mathcal{D}_{\text{target}}\bigr)$ indicates that the aggregated fact representations preserve the accuracy of the full documents, which is our desired outcome.

\section{Instruction Conversion Templates}
\label{app:sft_conversion}

As explained in \cref{sec:methodology}, we convert the raw extracted facts into instruction-response pairs using a rule-based approach. For each fact, we randomly sample a template from a manually curated list of 25 realistic conversational instructions. These templates are shown in \cref{fig:sft_templates}.

\begin{figure}[!h]
\centering
\begin{tcolorbox}
\fbox{SFT Conversion Templates}\par \vspace{1mm}
\begin{verbatim}
[
    "Tell me something about {entity}",
    "What fact can you tell me about {entity}?",
    "Q: What is a fact about {entity}?",
    "### Question\nWhat is a fact you can tell me about {entity}?\n",
    "### Question:\nWhat is a fact you can tell me about {entity}?\n",
    "Give me a fact about {entity}.",
    "What is a fact about {entity}?",
    "Tell me a fact about {entity}",
    "Share a single fact about {entity}.",
    "Can you give me one fact about {entity}?",
    "Provide a fact about {entity}.",
    "What’s a quick fact about {entity}?",
    "Do you have one fact about {entity}?",
    "What’s an interesting fact about {entity}?",
    "Tell me a quick fact about {entity}",
    "What’s a single detail about {entity}?",
    "What’s one thing you know about {entity}?",
    "What’s one piece of information about {entity}?",
    "What’s one fact you know about {entity}?",
    "Tell me one fact about {entity}.",
    "What’s one fact about {entity}?",
    "Here's a fact about {entity}.",
    "I want to know a fact about {entity}.",
    "### Question\nWhat is a fact about {entity}?\n### Answer\n",
    "### Question\nCan you tell me a fact about {entity}?",
]
\end{verbatim}
\end{tcolorbox}
\caption{A full list of all rule-based templates used to conver raw facts into SFT-compatible samples..}
\label{fig:sft_templates}
\end{figure}

\section{Prompts}
\label{app:prompts}
To encourage experimentation and further research, we release the full list of prompts used in this work. The prompts used in Knowledge-Instruct are shown in~\cref{fig:knowledge_entities,fig:knowledge_factoids,fig:knowledge_context}, while the rephrasing prompts used for Rephrase CPT are shown in~\cref{fig:rephrase_system,fig:rephrase_prompts_1,fig:rephrase_prompts_2,fig:rephrase_prompts_3}.

\begin{figure}[!h]
\centering
\begin{tcolorbox}
\fbox{Entity Extraction Prompt}\par \vspace{1mm}
Extract ALL entities from the following text.
You must be extremely thorough and ensure that you do not miss any entities.
You should aim to extract even the most obscure entities that may be mentioned in the text.
Look even for entities that might seem trivial or insignificant.

\vspace{5mm}
\fbox{Entity Extraction Follow-up Prompt}\par \vspace{1mm}
You have already extracted some entities. However, it appears you may have missed certain entities in the text. 

\textbf{Task:}
\begin{enumerate}
    \item Re-check the original text carefully.
    \item If you discover any additional entities not in your previously provided list, add them to a new list.
    \item Make sure you do not repeat any previously extracted entities.
    \item Again, aim to be as thorough as possible and look for even the most obscure/trivial entities.
\end{enumerate}

\textbf{Output Format:}
\begin{itemize}
    \item Return only the new entities.
    \item If you are absolutely confident there are no additional entities, return an empty JSON array [].
\end{itemize}

\textbf{Important:}
\begin{itemize}
    \item Only return newly discovered entities.
    \item Do not repeat any entities you have already extracted in previous responses in our conversation.
    \item You must be absolutely sure no entities are missing to return an empty array.
\end{itemize}

\end{tcolorbox}
\caption{Entity extraction prompts.}
\label{fig:knowledge_entities}
\end{figure}

\begin{figure*}[!h]
\centering
\begin{tcolorbox}
\fbox{Fact Extraction Prompt}\par \vspace{1mm}
You are given a document and an entity. 
Your task is to extract and list all factual statements about the given entity directly from the document. 
The entity is \texttt{\{entity\}}, and the document is given below. 
Follow these requirements:

\begin{enumerate}
    \item Use only information explicitly stated in the document. Do not infer or assume any details not supported by the text.
    \item Present your answer as a list.
    \item Each factual statement must:
    \begin{itemize}
        \item Be a complete, standalone sentence that clearly states a fact about the entity.
        \item Mention the entity by its full name exactly as given in the 'Entity' section.
        \item Provide enough context so it can be understood on its own.
    \end{itemize}
    \item Include only facts that are directly related to the given entity and explicitly mention it.
    \begin{itemize}
        \item Do not include any facts that do not directly reference the entity by name.
        \item Do not include any facts about other entities unless they are directly connected to the entity.
    \end{itemize}
    \item Importantly, if the entity is a numerical value, make sure to extract only facts that directly include the specific numerical value, and not general statements about numbers or quantities. 
    If a fact does not include the specific numerical value, it should not be included.
\end{enumerate}

\vspace{5mm}
\fbox{Fact Extraction Follow-up Prompt}\par \vspace{1mm}
Excellent! However, it is possible that the document contains more information about the entity than what you have listed.
Now, go over the document again, and search for any additional facts about the entity that you may have missed.
You must closely follow all the requirements mentioned in the previous prompt.
Specifically, ensure that each fact is directly related to the entity and explicitly mentions it by name.

If you found any new facts, return a list with just the new facts and nothing else.
If no additional facts about the entity can be found, just return an empty list.
\end{tcolorbox}
\caption{Fact extraction prompts.}
\label{fig:knowledge_factoids}
\end{figure*}

\begin{figure*}[!htb]
\centering
\begin{tcolorbox}
\fbox{Contextualize Prompt}\par \vspace{1mm}
You are provided with a factual statement about the following subject: '\{subject\}'.
Specifically, the factual statement focuses on the following entity: '\{entity\}' within the context of the subject.
If the statement does not explicitly mention the subject or entity full names and you believe that including them would provide additional clarity, you should include them in a way that fits naturally within the sentence.
Else, if the statement is already clear, mentions both the subject and entity in their full names, and does not require any additional context, you can leave it as is and return the statement as it is.
Return just the contextualized statement and nothing else.

\vspace{5mm}
\fbox{Verify Context Prompt}\par \vspace{1mm}
You have two inputs:
\begin{enumerate}
    \item An \textbf{entity} (e.g., a name, place, or keyword).
    \item A \textbf{statement} (one or more sentences).
\end{enumerate}

Your task is to determine whether the entity appears \textbf{in any form} within the statement. This includes:
\begin{itemize}
    \item Exact matches (e.g., the entity as-is).
    \item Variations of the entity (e.g., case differences, plural vs. singular).
    \item Recognizable references or abbreviations that clearly refer to the entity.
\end{itemize}

\textbf{Output Rules:}
\begin{itemize}
    \item If the entity appears, you must output <verdict>Yes</verdict>.
    \item If the entity does not appear at all, you must output <verdict>No</verdict>.
    \item You are allowed to describe your thinking process before providing the final verdict in the specified format.
\end{itemize}

\textbf{Entity:}
\texttt{'{entity}'}

\textbf{Statement}
\texttt{'{fact}'}

\end{tcolorbox}
\caption{Contextualization prompts.}
\label{fig:knowledge_context}
\end{figure*}

\begin{figure*}[!h]
\centering
\begin{tcolorbox}
\fbox{Rephrase System Prompt}\par \vspace{1mm}
You are an expert in text modification and paraphrasing.

Your task: 
\begin{itemize}
    \item You will be given an input text (below).
    \item You must produce \{n\_para\} distinct, long paraphrased versions of that text.
\end{itemize}

Requirements:
\begin{enumerate}
    \item \textbf{Retain Meaning \& Facts:} Each paraphrase must preserve the original text’s meaning, factual accuracy, and all specific details. Do not remove, alter, or add any factual information.
    \item \textbf{Variety in Paraphrasing:} Each paraphrase should be substantially different from both the original and from each other in terms of vocabulary, sentence structure, and style.
    \item \textbf{Maintain Length:} Each paraphrase should be approximately the same length as the original text.
    \item \textbf{No Additions or Omissions:} Do not introduce external information or assumptions not present in the original text. Do not remove any facts present in the original.
    \item \textbf{Preserve Specifics:} Do not change proper names, titles, dates, numbers, locations, direct quotes, or other specific references. 
    \item \textbf{Clarity and Tone:} Maintain the original clarity and intended tone. The result should read naturally and coherently.
    \item \textbf{Output Format:} Return the results as a Python dictionary with keys 'paraphrase\_1', ..., 'paraphrase\_{\{n\_para\}}'. Each value should be a single string containing the full paraphrased text.
    \item \textbf{Include Source:} If the source of the text is mentioned in the input (e.g., title, author, publication, etc.), you must include it in the paraphrased versions.
\end{enumerate}

\end{tcolorbox}
\caption{Rephrasing system prompt.}
\label{fig:rephrase_system}
\end{figure*}

\begin{figure*}[!h]
\centering
    \begin{tcolorbox}
    \fbox{Reshuffle Prompt}\par \vspace{1mm}
    Your sub-task is to rearrange the order of the sentences in the text provided below to provide a long paraphrased version.
    You must significantly change the sequence of sentences, ensuring the final text follows a clear, coherent, and logical structure.
    While reshuffling, you can make minor modifications to the wording to enhance the flow and coherence of the text.
    Your output must adhere to the previously mentioned requirements given in the system prompt.
    
    \vspace{5mm}
    \fbox{Reword Prompt}\par \vspace{1mm}
    Your sub-task is to rephrase the text provided below, focusing on the choice of words and sentence structure.
    Your goal is to replace as many words and phrases as possible with synonyms or alternative expressions while maintaining the original meaning, facts, and details.
    You must ensure that the rephrased text reads naturally and coherently.
    Your output must adhere to the previously mentioned requirements given in the system prompt. 
    
    \vspace{5mm}
    \fbox{Restructure Prompt}\par \vspace{1mm}
    Your sub-task is to rewrite the text provided below, focusing on the structure of the sentences.
    You should rephrase the text by changing the sentence structures, altering the word order, and varying the length of sentences while preserving the original meaning, facts, and details.
    You must ensure that the rephrased text reads naturally and coherently.
    Your output must adhere to the previously mentioned requirements given in the system prompt.
    
    \vspace{5mm}
    \fbox{Tense Variation Prompt}\par \vspace{1mm}
    Your sub-task is to subtly adjust the verb tenses and aspects in the text provided below.
    Where appropriate, change some simple past tenses to past perfect, or present tenses to present continuous, while still accurately reflecting the same time frames.
    Do not introduce new factual content, maintain the same approximate length, and preserve the original meaning and details.
    Your output must adhere to the previously mentioned requirements given in the system prompt.
    
    \vspace{5mm}
    \fbox{Invert Prompt}\par \vspace{1mm}
    Your sub-task is to produce a long paraphrased version of the provided text by inverting its structure. Specifically, you must:
    \begin{itemize}
        \item Reverse the overall order of the sentences, starting from the end of the original text and working backward toward the beginning.
        \item While inverting, you may make adjustments to the wording and sentence boundaries to ensure a coherent, logical flow.
        \item Preserve all factual information, names, dates, and other specifics without adding, removing, or distorting any facts.
        \item Maintain the approximate length of the original text and reflect its general tone and clarity.
        \item Ensure the final output reads naturally, as though the text was originally structured in this inverted order.
    \end{itemize}
    
    Your output must adhere to the requirements outlined in the system prompt.
    \end{tcolorbox}
\caption{Rephrasing prompts.}
\label{fig:rephrase_prompts_1}
\end{figure*}

\begin{figure*}[!h]
\centering
    \begin{tcolorbox}
    \fbox{Summarization Prompt}\par \vspace{1mm}
    Your task is to produce a comprehensive and highly detailed knowledge-focused summary of the provided text. \\
    Requirements:
    \begin{enumerate}
        \item \textbf{Thoroughness and Accuracy:} Include all factual information, key points, and essential details, ensuring nothing of importance is omitted.
        \item \textbf{No Alteration of Facts:} The summary must faithfully reflect the original text without adding, removing, or distorting any facts.
        \item \textbf{No External Content:} Do not introduce assumptions, opinions, or details not present in the original text.
        \item \textbf{Lexical Variety:} Use a diverse range of vocabulary and phrasing to make the summary more engaging, while maintaining accuracy and avoiding unnecessary repetition.
        \item \textbf{Preserve Tone:} Reflect the general tone and style of the original text.
    \end{enumerate}
    
    Your goal is to create an extensive summary that captures the full breadth of the source material while maintaining clarity, factual integrity, and coherent organization.
    
    \vspace{5mm}
    \fbox{Detail Emphasis Prompt}\par \vspace{1mm}
    Your sub-task is to carefully review the provided text and produce a paraphrased version that highlights all minor details and subtle nuances that might otherwise be overlooked.
    
    \textbf{Requirements:}
    \begin{enumerate}
        \item \textbf{Thorough Attention to Detail:} Identify and preserve every small fact, reference, or subtle hint present in the original text, ensuring nothing is lost.
        \item \textbf{No Alteration of Facts:} Accurately reflect every piece of information without distorting or omitting any detail.
        \item \textbf{Clarity and Coherence:} Present the paraphrased text clearly, making sure that the small details fit naturally into the narrative and contribute to overall coherence.
        \item \textbf{Fidelity to Original Tone:} Maintain the original tone, length, and style, incorporating all subtle elements in a way that feels organic and readable.
    \end{enumerate}
    
    Your goal is to produce a paraphrased version that gives as much importance to minor aspects as it does to major points, ensuring full fidelity to the original text.
    
    \end{tcolorbox}
\caption{Rephrasing prompts.}
\label{fig:rephrase_prompts_2}
\end{figure*}

\begin{figure*}[!h]
\centering
\begin{tcolorbox}
\fbox{Middle Restructure Prompt}\par \vspace{1mm}
Your sub-task is to produce a long paraphrased version of the provided text by reorganizing its structure to begin from its midpoint.

\textbf{Guidelines:}
\begin{enumerate}
    \item Identify a clear midpoint in the text and start your rewriting from that point.
    \item Move forward through the latter half of the text after this midpoint, maintaining logical flow and coherence.
    \item Once you reach the end of the original text, continue by incorporating the initial portion (the beginning section) at the end, so that the sequence now runs from the middle to the end, and then from the start to the middle.
    \item While restructuring, you may slightly modify wording and sentence boundaries to enhance clarity, coherence, and readability.
    \item Preserve all factual information and details. Do not add or remove any facts. Names, dates, locations, and other specifics must remain accurate.
    \item Reflect the original tone and style.
\end{enumerate}

Your output must adhere to the requirements outlined in the system prompt.

\vspace{5mm}
\fbox{Simplify Prompt}\par \vspace{1mm}
Your task is to simplify the provided text by using more common language, and clearer expressions.
You must ensure that the rephrased text reads naturally and coherently without losing any factual information or details.
Your output must adhere to the requirements outlined in the system prompt.
\end{tcolorbox}
\caption{Rephrasing prompts.}
\label{fig:rephrase_prompts_3}
\end{figure*}

\begin{figure*}[!htb]
\centering
\begin{tcolorbox}
\fbox{Fictional Company Profile}\par \vspace{1mm}
I want to create a dataset of fictional companies. Each company should resemble real companies in general, but not mimic any specific company directly.

Let's start with \texttt{\{company\}}. Provide a comprehensive company profile, similar to what you might find in a detailed Wikipedia article about a major corporation.

The profile should include:

Company history and background (e.g., founders, founding year, and early milestones). 
Key financial highlights (e.g., market capitalization, revenue, major funding rounds, or IPO details). 
A detailed description of its products, services, or business model, including major innovations and failures. 
Current corporate structure, key executives, and employee headcount. 
Offices, headquarters, and geographical reach. 
Recent controversies, challenges, or notable news (e.g., lawsuits, scandals, failed products, or market competition). 
Details about its industry standing, competitors, and long-term strategic goals. 
Write the response as detailed, paragraphed, and well-structured text. Use an informative, neutral tone similar to that of a Wikipedia article. Incorporate small, specific details to make the profile feel authentic and nuanced.

\end{tcolorbox}
\caption{Fictional company profile generation prompt.}
\label{fig:company_profile}
\end{figure*}

\begin{figure*}[!htb]
\centering
\begin{tcolorbox}
\fbox{Fictional Company Product Catalog}\par \vspace{1mm}
Based on the company profile provided, create a detailed product catalog for \texttt{\{company + "\textbackslash n\textbackslash n" + profile\}}. Ensure it is as rich and detailed as a real-world product catalog. 
\\
The catalog should include:

At least 20 products, spanning multiple categories that align with the company's business model. 
Full product descriptions, written with a mix of marketing language and technical details. 
Comprehensive specifications for each product, including features, technical specs, and any unique selling points. 
Pricing details (e.g., MSRP, subscription fees, or tiers for different customer segments). 
Any notable reviews, industry recognition, or customer use cases for certain products. 
Notes on availability (e.g., discontinued products, regional exclusivity, or supply issues). 
Write this in unstructured, free-form text. Think of how a large company might present its offerings in an online store, brochure, or detailed catalog. Ensure the tone is professional but engaging.

\end{tcolorbox}
\caption{Fictional company product catalog generation prompt.}
\label{fig:company_catalog}
\end{figure*}

\begin{figure*}[!htb]
\centering
\begin{tcolorbox}
\fbox{Fictional Company Financial Report}\par \vspace{1mm}
Recall the company profile and product catalog for \texttt{\{company + "\textbackslash n\textbackslash n" + profile + "\textbackslash n\textbackslash n" + catalog\}}. Create a full financial report of its performance for the past quarter. The report should be consistent with the previous information provided and sound realistic from an accounting and corporate reporting perspective.

The financial report should include:

Revenue, gross profit, and net income figures, broken down by business segments (e.g., product lines, geographical regions, or divisions). 
Key financial ratios (e.g., profit margins, debt-to-equity ratio, etc.). 
Year-over-year (YoY) or quarter-over-quarter (QoQ) comparisons, highlighting trends. 
Major revenue drivers and cost challenges, including any standout successes or failures. 
Commentary on the company's cash flow, liquidity, and capital expenditures. 
Updates on stock performance, shareholder returns, or dividend announcements (if applicable). 
Notable financial risks or opportunities (e.g., lawsuits, declining market share, or upcoming product launches). 
Write this in detailed, unstructured free-form text, as if prepared by the finance department of a major corporation for internal or investor reporting. Include realistic terminology and phrasing to create an authentic tone.

\end{tcolorbox}
\caption{Fictional company financial report generation prompt.}
\label{fig:company_financial_report}
\end{figure*}

\FloatBarrier
\section{Data Samples}
\label{app:data_samples}
To provide further clarity regarding Knowledge-Instruct, we provide an example of the method and dataset for a specific example from the Companies dataset: 
\begin{quote}
    QuantumQuirkySprouts Labs: An agricultural biotech startup founded by a retired quantum physicist and her vegan daughter.
\end{quote} 
We include here examples of facts generated for the company "QuantumQuirkySprouts Labs", as well as questions about the company from the Companies benchmark.~\cref{fig:wiki_overview} shows an example of a company's 'wiki overview',~\cref{fig:sample_facts} shows examples of extracted facts, and~\cref{fig:questions_sample} show sample questions about the company.

\begin{figure}[!h]
\centering
\begin{tcolorbox}
\fbox{Context: QuantumQuirkySprouts Labs}\par \vspace{1mm}

\textbf{Company History and Background}  

QuantumQuirkySprouts Labs is an innovative agricultural biotech startup founded in 2015 by Dr. Eleanor Hawthorne, a retired quantum physicist, and her daughter, Mia Hawthorne, a passionate vegan and environmental advocate. The company was established in Palo Alto, California, with the vision of integrating advanced quantum mechanics principles into sustainable agriculture practices. The unique combination of Dr. Hawthorne's expertise in quantum physics and Mia's knowledge of plant-based nutrition laid the foundation for a company that aims to revolutionize how food is grown and consumed.

\textbf{Key Financial Highlights}  

QuantumQuirkySprouts Labs has experienced robust financial growth since its inception. By 2022, the company had secured a Series C funding round worth \$75 million, led by GreenFuture Capital and EcoInnovate Ventures. These investments have propelled the company's market capitalization to approximately \$400 million. Despite its rapid growth, QuantumQuirkySprouts has yet to go public, with a potential IPO projected for the late 2020s as part of its long-term strategic plan.

\textbf{Corporate Structure and Key Executives}  

QuantumQuirkySprouts Labs is structured as a private company, with Dr. Eleanor Hawthorne serving as the Chief Executive Officer and Mia Hawthorne as the Chief Innovation Officer. The executive team includes industry veterans such as Victor Lin, the Chief Technology Officer, and Sara Patel, the Chief Financial Officer. As of 2023, the company employs approximately 250 people, including a diverse team of scientists, engineers, and business professionals committed to advancing sustainable agriculture.

\textbf{Offices, Headquarters, and Geographical Reach}  

The company's headquarters remain in Palo Alto, California, with additional research facilities located in Austin, Texas, and Cambridge, Massachusetts. QuantumQuirkySprouts has expanded its geographical reach to include pilot projects and partnerships in Europe and Asia, focusing on regions with burgeoning interest in sustainable agricultural technologies.

\end{tcolorbox}
\caption{Wiki overview of QuantumQuirkySprouts Labs.}
\label{fig:wiki_overview}
\end{figure}

\begin{figure}[!h]
\centering
\begin{tcolorbox}
\fbox{Sample Facts}\par \vspace{1mm}

\begin{itemize}
    \item \textbf{QuantumQuirkySprouts Labs}: QuantumQuirkySprouts Labs is an agricultural biotech startup founded in 2015 by Dr. Eleanor Hawthorne and her daughter, Mia Hawthorne.
    \item \textbf{QuantumQuirkySprouts Labs}: QuantumQuirkySprouts Labs was established in Palo Alto, California.
    \item \textbf{Dr. Eleanor Hawthorne}: Dr. Eleanor Hawthorne, a retired quantum physicist at QuantumQuirkySprouts Labs, is known for her contributions to the field.
    \item \textbf{Dr. Eleanor Hawthorne}: Dr. Eleanor Hawthorne co-founded QuantumQuirkySprouts Labs in 2015.
    \item \textbf{Dr. Eleanor Hawthorne}: Dr. Eleanor Hawthorne founded QuantumQuirkySprouts Labs in Palo Alto, California.
    \item \textbf{Mia Hawthorne}: Mia Hawthorne is the daughter of Dr. Eleanor Hawthorne, a retired quantum physicist associated with QuantumQuirkySprouts Labs.
    \item \textbf{Mia Hawthorne}: Mia Hawthorne, a key figure at QuantumQuirkySprouts Labs, is a passionate vegan and environmental advocate.
    \item \textbf{Mia Hawthorne}: Mia Hawthorne serves as the Chief Innovation Officer at QuantumQuirkySprouts Labs.
\end{itemize}

\end{tcolorbox}
\caption{Example of facts generated.}
\label{fig:sample_facts}
\end{figure}

\begin{figure}[!h]
\centering
\begin{tcolorbox}
\fbox{Questions}\par \vspace{1mm}

\begin{itemize}
    \item \textbf{Question}: The question refers to the company "QuantumQuirkySprouts Labs". Who are the founders of QuantumQuirkySprouts Labs, and what are their backgrounds?  \\
          \textbf{Answer}: Dr. Eleanor Hawthorne, a retired quantum physicist, and her daughter, Mia Hawthorne, a passionate vegan and environmental advocate.
    \item \textbf{Question}: The question refers to the company "QuantumQuirkySprouts Labs". What position does Mia Hawthorne hold at QuantumQuirkySprouts Labs?  \\
          \textbf{Answer}: Chief Innovation Officer.
\end{itemize}

\end{tcolorbox}
\caption{Sample question-answer pairs for QuantumQuirkySprouts Labs.}
\label{fig:questions_sample}
\end{figure}

\end{document}